\documentclass[conference]{IEEEtran}
\IEEEoverridecommandlockouts
\usepackage{cite}
\usepackage{amsmath,amssymb,amsfonts}
\usepackage{algorithm}
\usepackage{algorithmic}
\usepackage{graphicx}
\usepackage{textcomp}
\usepackage{xcolor}
\usepackage{subfigure}
\usepackage{booktabs}
\usepackage{geometry}
\usepackage{multirow}
\usepackage{url}
\def\BibTeX{{\rm B\kern-.05em{\sc i\kern-.025em b}\kern-.08em
    T\kern-.1667em\lower.7ex\hbox{E}\kern-.125emX}}
\begin{document}
\columnsep 0.25in

\title{A Novel Optimized Asynchronous Federated Learning Framework 
\thanks{This work was supported by the Key Research and Development Program of Hainan Province (Grant No.ZDYF2020040), Major science and technology project of Hainan Province(Grant No.ZDKJ2020012), Hainan Provincial Natural Science Foundation of China (Grant Nos. 2019RC098), National Natural Science Foundation of China (NSFC)(Grant No.62162022, 62162024 and 61762033) and Key Projects of Innovation and Entrepreneurship for Undergraduates of Hainan University (Grant No.20210102).
}
}

\author{
    \IEEEauthorblockN{Zhicheng Zhou$^{1,3}$, Hailong Chen$^{1,3}$, Kunhua Li$^{2,3}$, Fei Hu$^{3}$, Bingjie Yan$^{1,3,*}$, Jieren Cheng$^{1,4,*}$ \\
    Xuyan Wei$^{2,3}$, Bernie Liu$^{3}$, Xiulai Li$^{1}$, Fuwen Chen$^{3}$ and Yongji Sui$^{3}$}
    \IEEEauthorblockA{$^1$ School of Computer Science and Technology, Hainan University, Haikou, China}
    \IEEEauthorblockA{$^2$ School of Cyberspace Security(School of Cryptology)
 , Hainan University, Haikou, China}
    \IEEEauthorblockA{$^3$ RobAI-Lab, Hainan University}
    \IEEEauthorblockA{$^4$ Hainan Blockchain Technology Engineering Research Center, Haikou, China}
    \IEEEauthorblockA{\{zc.zhou, bj.yan\}@ieee.org, \{cjr22\}@163.com}
}
\maketitle

\begin{abstract}
Federated Learning (FL) since proposed has been applied in many fields, such as credit assessment, medical, etc. Because of the difference in the network or computing resource, the clients may not update their gradients at the same time that may take a lot of time to wait or idle. That's why Asynchronous Federated Learning (AFL) method is needed. The main bottleneck in AFL is communication. How to find a balance between the model performance and the communication cost is a challenge in AFL. This paper proposed a novel AFL framework VAFL. And we verified the performance of the algorithm through sufficient experiments. The experiments show that VAFL can reduce the communication times about 51.02\% with 48.23\% average communication compression rate and allow the model to be converged faster. The code is available at \url{https://github.com/RobAI-Lab/VAFL}
\end{abstract}


\section{Introduction}

Machine learning and artificial intelligence need to be fed lots of data to get a well-performed model, which cased the demand of data that is stored in every device. For instance, smart city\cite{zheng2021applications}, medical\cite{yan2021experiments}, and credit assessment\cite{yang2019federated}. But, this data is sensitive. At the same time, the relationship between data is often isolated and fragmented. This limits its application for processing data. 

In this context, federated learning was proposed. Federated learning is a privacy-preserving distributed machine learning technique. The data holders only need to provide a local model which was trained using their local data to interact with the central server. In this paradigm, only gradients need to be transmitted rather than the raw data, which can effectively reduce privacy risks. In federated learning, the client downloads the global model from the central server and trains the local data under the coordination of the central server. After a round of training is completed, the trained local model is uploaded to the central server for weighted aggregation to complete the update. After some iterations, the performance of the model is similar to the results of centralized machine learning. However, there are some challenges in federated learning\cite{yang2019federated}\cite{chen2019communication}. Traditional federated learning has obvious shortcomings in flexibility and scalability. These challenges are shown below:

\begin{itemize}
    \item Heterogeneous device: Since training data is stored on the client-side, heterogeneous device client used to have different data sizes, computing power, network, etc., and are affected by unstable communication conditions, the efficiency of traditional synchronous federated learning can be affected by the presence of dropped users. For example, when a few clients are disconnected due to network problems, other clients and server have to wait for them, causing the federated learning process slowly and spend lots of time idle.
    \item Heterogeneous data: Since federated learning data are only stored locally, the inconsistency of this distribution among data in practical applications leads to significant degradation of model performance, such as feature distribution skew, label distribution skew, quantity skew\cite{li2021federated,kairouz2019advances}.
    \item Security risk: Even though federated learning does not involve the client raw data in the transmission, there is still a considerable security risk. For example, attackers may use GAN to learn potential information in the gradients\cite{2017Deep}.

\end{itemize}

\begin{figure*}[tbp]
\centerline{\includegraphics[width=0.6\linewidth]{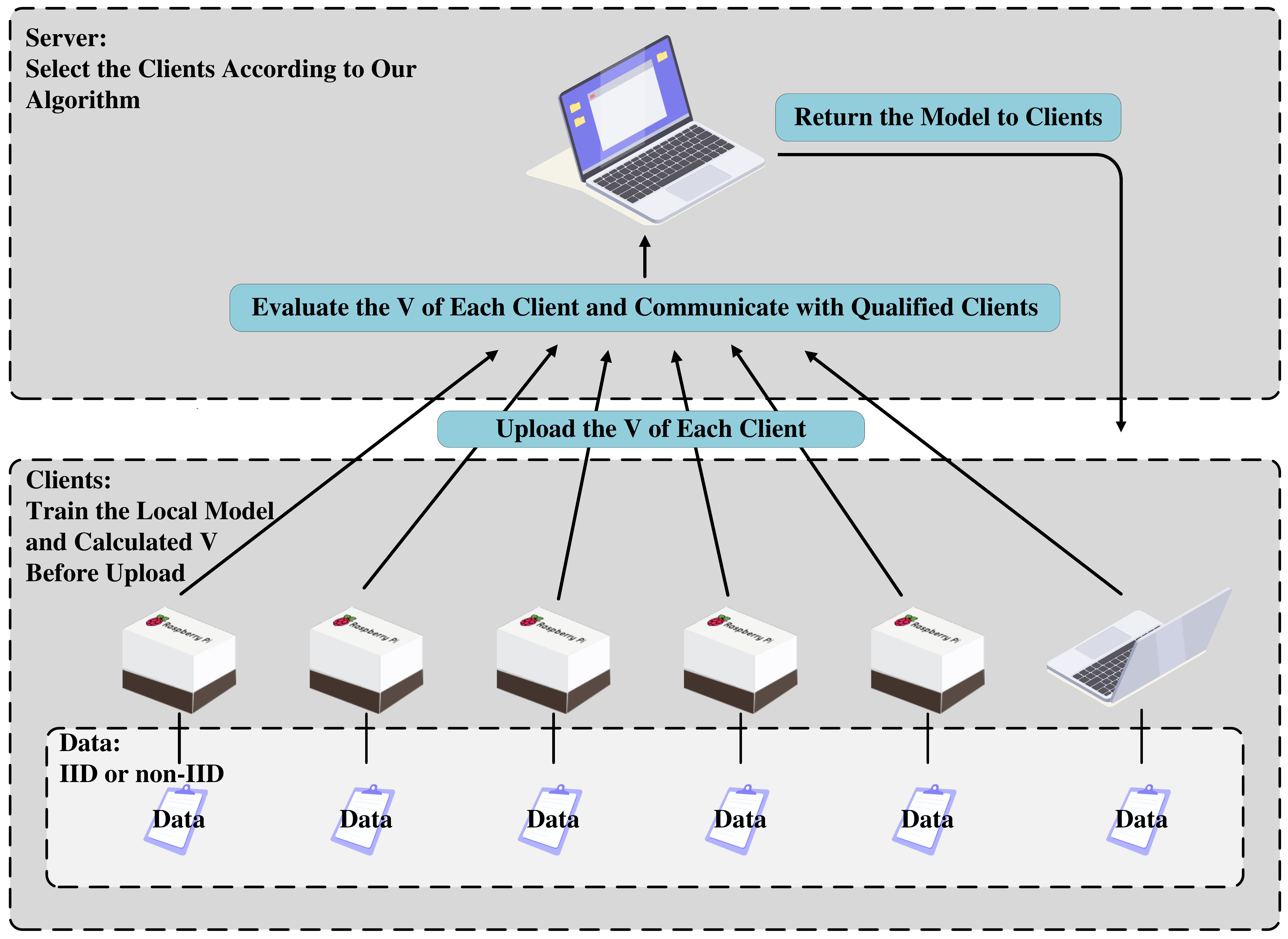}}
\caption{Asynchronous federated learning framework with VAFL.}
\label{VAFL}
\end{figure*}

In order to address these issues and challenges, FedAvg\cite{mcmahan2017communication}, FedProx\cite{li2018federated}, SkewScout\cite{hsieh2020non}, and other algorithms have been proposed one after another. Besides, asynchronous federated learning was proposed. Asynchronous federated learning mainly addresses the problem of how the central server parameters should be aggregated in the presence of any heterogeneous clients (heterogeneous data distribution, computing resources, network conditions, etc.).

In this paper, we propose a new asynchronous federated learning optimization algorithm (VAFL) for the above challenges (mainly the first two). To address the first challenge, we consider the number of participating clients and provide an improved method for judging clients by checking obsolescence, introducing a function for evaluating the communication value of clients so that random selection is no longer performed. By selecting high-quality clients, the global model accuracy is improved. For the second challenge, we make a division between IID and Non-IID on the datasets, aiming to cope with practical federated learning scenarios, and explore applications in Non-IID environments.

In summary, our contributions are as follows:

\begin{itemize}
    \item We proposed a novel asynchronous federated learning algorithm VAFL, which can reduce the communication cost and the idle time and make the model converge faster.
    \item We set up evaluation formulas for client selection to determine which clients upload the model. Experiment results show that this method accelerates the accuracy obtained at the beginning of training, and it also ensures that the communication compression rate can be improved to a certain extent.
    \item We perform extensive experiments and validate the performance of VAFL.
\end{itemize}



\section{Related work}

In recent years, the Federated Learning framework proposed by Google has attracted great attention from academic scholars. The standard federated learning model needs to learn a single global statistical model from the stored data of a large number of remote devices. However, the standard model faces challenges such as high communication costs, system heterogeneity, statistical heterogeneity, and privacy issues\cite{li2020federated}.

In response to these challenges, McMahan\cite{mcmahan2017communication}proposed a method for synchronous training of the Federated learning Average Algorithm (FedAvg), and recently Q Li\cite{li2021model} proposed MOON based on FedAvg. These algorithms are updated locally. The cost of communication is reduced; not only that, privacy protection is based on synchronization operations such as differential privacy\cite{bhowmick2018protection}and secure aggregation\cite{bonawitz2017practical} can solve the privacy problem to a certain extent.

However, many types of research on federated learning today are mainly about Non-IID data sets and data imbalances, and most of the researches are based on synchronization algorithms. However, federated learning based on synchronization algorithms cannot solve the problem of excessive communication costs. Issues such as differences in terminal performance due to different equipment terminals.

In response to the above challenges, many researchers have begun to try asynchronous algorithms for federated learning. For example, EAFLM proposed by Lu\cite{lu2020asynchronous}, this method can achieve the goal of reducing communication costs without searching for the optimal solution; SS Diwangkara\cite{diwangkara2020study}implemented it through an asynchronous aggregation algorithm The rapid convergence of each node and the reduction of the update frequency of the parameters can not only reduce the communication cost, but the method also achieves good results in the Non-IID data set. Although the above methods can reduce the communication cost better, they did not consider the serious delay caused by the lagging party. For this reason, McMahan B\cite{mcmahan2017communication} proposed the use of multiple SGD updates and batch processing clients to alleviate the delay. Problem; Subsequently, Nishio T\cite{nishio2019client} suggested using the deadline method to solve the customer selection problem. Although this method can filter out some slow-response customers, it did not consider how to solve the model accuracy rate caused by the backwardness in training. Too low customers, for this reason, Chai\cite{chai2020tifl} proposed a hierarchical-based federated learning system (TIFL), which can divide customers into different levels according to the user's response delay in the federated system. To select customers based on the system, the system can not only solve the problem of the user falling behind but also further alleviate the problem of Non-IID data. However, the TIFL system adaptively adjusts the client's participation based on the response delay of the client's machine, and it cannot solve some clients with high participation but low data quality.

Aiming at the excessively high communication cost in the federated learning training and the lagging customers, we propose an optimized federated learning asynchronous algorithm based on the communication value calculation of the gradient, the number of clients, and the accuracy of the model(VAFL). Different from the method of Chai\cite{chai2020tifl}, we use the client's gradient change, the accuracy rate on the test set, and the number of clients to evaluate the client's communication value, thereby determining whether the client can continue to participate in federated optimization. At the same time, considering the problem of low performance of client devices in reality, when we set up the experimental platform, we used a Raspberry Pi with poor performance to simulate client devices.

\section{Methodology}

\subsection{Communication Value Calculation}
This section illustrates how VAFL is derived and worked. The communication value calculation is based on gradient, the number of clients, and model accuracy. The notations covered in this section are shown in Tab. \ref{tab1}.

\begin{table}[htbp]
	\caption{The Meaning of the Symbol}
	\label{tab1}
	\begin{center}
		\begin{tabular}{ccc}
			\toprule
			\textbf{Symbol}& \textbf{Meaning}\\
			\midrule 
			$V$& \text{Communication value of the client} \\ 
			$V_i$ & \text{The $i$th client's communication value} \\ 
			$\mathbf{\nabla}_i$ & \text{The $i$th client's gradient} \\ 
			$\mathbf{\nabla}_i^k$ & \text{Gradient at the $k$th round of training for the $i$th client} \\ 
			$Acc$ & \text{Accuracy of client models on the testset} \\ 
            $Acc_i$ & \text{Accuracy of the $i$th client model on the testset} \\ 
            $n$ & \text{Number of clients involved in federated learning} \\ 
			\bottomrule
		\end{tabular}
	\end{center}
\end{table}
We refer to the work of Chen et al.\cite{chen2020lasg}, where the main idea of Chen et al.'s work is that not all communication back and forth between the server and the clients is equally important. Selecting only the important clients for communication reduces the communication cost and also maintains the convergence speed. Instead, our idea is to let the more communication-valuable clients communicate with the server, which we define as $V$, where the communication value of the $i$th client is denoted as $V_i$.

How to determine the communication value of the client then becomes an important factor in the design of the algorithm. After following the prior work about the calculation of communication value, we decided to use the gradient of the clients' model as a variable in VAFL design. We use $\nabla_i$ to represent it. Considering that the model performance of each client reflects some extent the performance of the client and the combination of the datasets it has (in general, the more quality datasets it has, the better the performance of the model obtained by local training of the client will have better performance). For convenience, we choose an easy metric: the accuracy of the client model on the test set, as a variable for VAFL design. We define it as $Acc$, and The accuracy of the $i$th client model is ${Acc}_i$. Recently, it has also been demonstrated that special client selection methods can optimize the federated learning process and improve the performance of the final model\cite{2020TIFL}\cite{9237167}. Therefore, we choose the number of clients involved in federated learning as another variable in the design of VAFL. We define it as $N$;

Ultimately, we designed the evaluation formula for $V_i$:

\begin{equation}
V_i=\left \| \mathrm{\nabla}_i^{k-1}-\mathrm{\nabla}_i^k \right \| ^2 \times \left (1+\frac{N}{10^3}\right)^{Acc_{i}}
\label{Vi=}
\end{equation}

Where $\nabla_i^{k-1}-\nabla_i^k$ is the gradient difference of the model in the last two training processes of the $i$th client model. We use it to check whether the model is ``old''. This difference is smaller, indicating that the model is older, which means its $V$ is smaller; $n$ is the number of clients involved in federated learning, we divide it by ${10}^3$ and plus $1$ as the base of the power function with ${Acc}_i$ as the variable. Thus, as the number of clients involved in federated learning increases, the relative communication value of individual clients is further differentiated. The more valuable the client, the further the value will increase, while the less valuable the client, the further the value will decrease. To facilitate the design of VAFL.

\subsection{An Asynchronous Federated Optimization Algorithm}

This section introduce the design idea and algorithm flow of our asynchronous federated optimization algorithm based on client communication value.

We hope to compress the communication by measuring the communication value of each client participating in federated learning to decide whether they participate in federated optimization or not, so as to deal with the problem of communication bottleneck and unbalanced in asynchronous federated learning.

For the clients involved in federated learning, we calculate their communication value $V$ by using Equation \ref{Vi=}, and the server will be able to learn the V of each client. Add it to the asynchronous FedAvg only if the client's V satisfies the Formula \ref{Vi>=}.

\begin{equation}
V_i\geqslant \frac{\sum_{j=0}^{N}V_j}{N}
\label{Vi>=}
\end{equation}

The specific algorithm pseudo-code is shown in Algorithm \ref{[Algorithm 1]}

\begin{algorithm}
	\caption{V based Asynchronous FedAvg}
	\label{[Algorithm 1]}
    $N$ is the total number of clients , $\theta_t$ is the parameters of current model, $\theta_i^t$ is the parameters of current model with index $i$, $B$ is the batch size, $E$ is the number of local epochs, $S$ is a set of all clients model and $\eta$ is the learning rate. 
    \begin{algorithmic}[1]
        \STATE \emph{\textbf{Server Update}}
		\STATE Initialize $\theta_0$;
		\FOR{each round $t=1,2,\cdots$ }
        \FOR{each $i\in S$}
		\STATE $V_i \leftarrow$ ClientUpdate($i$, $\theta_i^t$);
		\STATE // clients update model on local device and upload the $V_i$ to server;
		\ENDFOR
		\STATE $\bar V = \frac{\sum_{i=0}^{N}V_i}{N}, \theta '=\varnothing $;
        \FOR{each $i\in S$}
		\IF{$V_i\geqslant \bar {V}$} 
        \STATE request $\theta_i^{t+1}$ from clients $i$;
        \STATE $\theta '$ append $\theta_i^{t+1}$;
		\ENDIF
		\ENDFOR
		\STATE $K\leftarrow$ the length of $\theta '$;
        \STATE $\theta^{t+1}\leftarrow$ $\sum_{i=1}^{K}\frac{n_i}{n}\theta_i^{t+1} $;
		\ENDFOR
		\STATE \emph{\textbf{Client Update}}
		\STATE $B \leftarrow$ Split user data into local mini-batch size $B$;
		\FOR{each local epoch $e$ from $1$ to $E$}
		\FOR{batch $b \in B$}
		\STATE $\theta\leftarrow \theta-\eta \bigtriangledown L(\theta;b)$;
		\ENDFOR
		\ENDFOR
		\STATE $V\leftarrow \left \| \mathrm{\nabla}^{k-1}-\mathrm{\nabla}^k \right \| ^2 * \left ( 1+\frac{N}{10^3}\right)^{Acc}$;
		\RETURN$V$ to server 
	\end{algorithmic}
\end{algorithm}

\section{Experiment Setting}
\label{sec:setting}

In this section, we will introduce how we built our experiment system and how VAFL is implemented. 

\subsection{Hardware Platform}
VAFL are designed for varying edge devices. Following this principle, we used five Raspberry Pi devices and two laptops to build our experiment system for federated learning for edge computing, which is more realistic, compared to the simulated asynchronous environment. Among them, the laptop as the central server is configured with a 6-core Intel(R) Core(TM) i7-9750H CPU running at 2.59GHz with 8GB of memory, and the rest of the devices as edge clients, where the laptop is configured with a 4-core Intel(R) Core(TM) i5-9300H CPU running at 2.40GHz and 8GB of memory, one Raspberry Pi device is Raspberry Pi 4B, which is configured with a 4-core ARM Cortex-A72 running at 1.50GHz and 4GB of memory, and the remaining four Raspberry Pi devices are Raspberry Pi 4B with 8GB of memory.

All devices are connected to the same LAN with a network bandwidth of 2.4GHz and the speed for receiving and transmitting are 216 Mbps and 120 Mbps.

\subsection{Framework}
We use PySyft\cite{ryffel2018generic} as the basic framework to build our asynchronous federated learning experiment system. PySyft is a Python library for Federated Learning, Differential Privacy, and Encrypted Computation. We use the client class and server class provided by PySyft for connecting via WebSocket to facilitate the construction of our federated learning system. We use PyTorch as the deep learning backend which is supported by PySyft.

We run our client program on each client (a total of five Raspberry Pi devices and one laptop, which cannot be informed of each other's existence and ensures a certain degree of security and privacy between the clients). Each client is connected to the server, which can only be informed about the model of each client and the total number of clients. The server uses VAFL to update the model, and then returns the model obtained by the algorithm to the client, and the client uses the model to predict. The architecture of our experiment system is shown in Fig. \ref{VAFL}.

\begin{figure}[htbp]
\centerline{\includegraphics[width=0.6\linewidth]{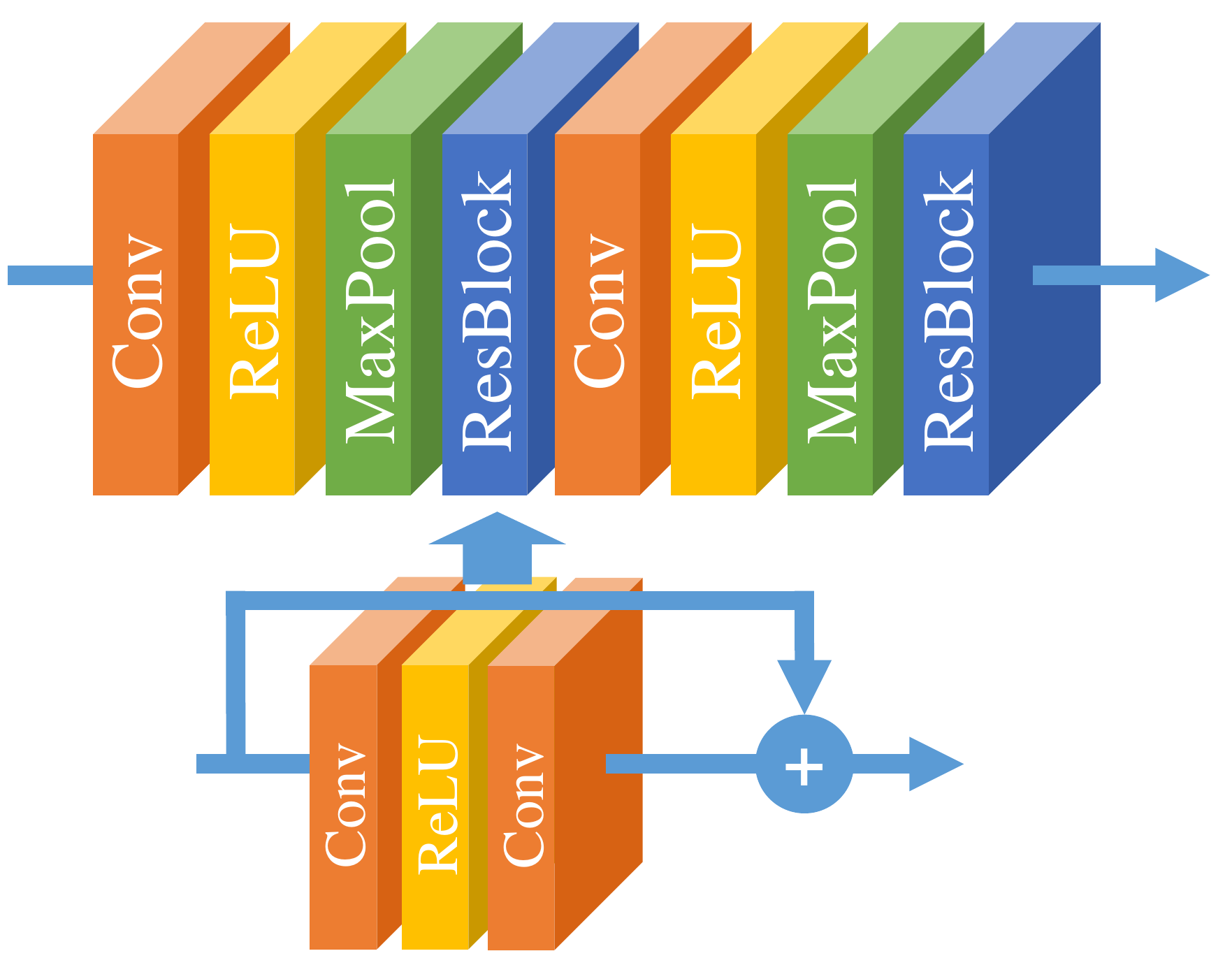}}
\caption{The ResNet structure we use.}
\label{ResNetS}
\end{figure}

We build our federated learning system using PySyft 0.2.4 with Python 3.7. We use ResNet\cite{2016Deep} as the core part of model based on PyTorch 1.4.0, and its network structure is schematically shown in Fig. \ref{ResNetS}.

\subsection{Data}

We constructed IID and Non-IID data distributed to each client using MNIST and conducted the experiments separately. For the IID case, the training set is equally distributed to all clients, each client contains samples with 10 labels. For the Non-IID dataset distribution, the training set varies in terms of labels and the number of samples with different labels. Some clients containing all labels and a large number of samples under each label, and some clients containing only a small number of labels and some samples under each label.

By constructing such a dataset, we hope to further explore the effect of whether the data is IID on the asynchronous federated learning system.

\subsection{EAFLM}
The main idea of EAFLM\cite{lu2020asynchronous} is to ignore the ``lazy'' nodes in rounds and only communicate with the ``hard-working'' nodes. We were inspired by this and improved, so we choose Lu's algorithm for comparison. The difference between VAFL and EAFLM is that EAFLM selected the parameters and gradient of the model as the main factors for the design of the algorithm, whose algorithm is mainly as follows.

\begin{equation}
\left \| \mathrm{\nabla}_i\left ( \theta^{k-1}\right ) \right \|^2\leqslant \frac{1}{\alpha ^2\beta m^2}\left \| \sum_{d=1}^{D}\xi_d\left ( \theta^{k-d}-\theta^{k-1-d} \right ) \right \|^2
\label{Vi>=2}
\end{equation}

Where $\nabla_i\left(\theta^{k-1}\right)$ is the $k-1$ th round gradient calculated by client $i$ based on the $k-1$ th round parameter. $\theta^{k-1}$ is the $k-1$th round parameter of the server. $\beta$, $\xi_d$ and $D$ are constant coefficients. $\alpha \epsilon \left ( 0,1 \right )$ is the adjustable parameter in the parameter weights, which determines the decay rate. As $\alpha$ increases, the decay rate of the parameter weights increases. In the experiments of this paper, $\xi_d=\frac{1}{D}$, $D=1$, $\alpha=0.98$.

\section{Experiment and Analysis}
In this section, we conduct four experiments, setting 3 or 7 clients with iid or non-iid data, to validate our proposed asynchronous federated optimization algorithm VAFL.

\subsection{Experiment Metrics}
We use the accuracy rate ($Acc$) and communication compression rate ($CCR$) as the evaluation metrics of the experiments.
Where $Acc$ is the highest $Acc$ rate in multiple experiments and the communication compression rate ($CCR$) is calculated as:

\begin{equation}
CCR= \frac{C_{t0}-C_{t1}}{C_{t0}}\times 100%
\label{CCR}
\end{equation}

Where $C_{t0}$ is the communication times before compression and $C_{t1}$ is the communication times after compression.

The communication compression rate reflects how much the communication between the clients and the server is compressed, and the larger the compression rate, the higher the degree of compression. Although excessive compression rate reduces communication loss and saves time, it is usually accompanied by a decrease in model $Acc$.

Our experiments use two metrics to evaluate the effectiveness of VAFL, which is to achieve a certain communication compression rate while ensuring the loss of model $Acc$. The parameters for each of VAFL are shown in Tab. \ref{tab2}.

\begin{table}[htbp]
	\caption{Parameters}
	\label{tab2}
	\begin{center}
		\begin{tabular}{ccc}
			\toprule
			\textbf{Symbol}& \textbf{Meaning}& \textbf{Value} \\
			\midrule 
			$r$& \textit{Local Training Rounds} & $5$ \\ 
			$E$ & \textit{The number of local epoch} & $1$ \\ 
			$B$ & \textit{Batch Size} & $32$ \\ 
			$\eta$ & \textit{Learning Rate} & $0.1$ \\ 
			$R$ & \textit{Total Training Rounds} & $200$ \\ 
			\bottomrule
		\end{tabular}
	\end{center}
\end{table}

\subsection{Experiment}

We conduct four experiments, setting 3 or 7 clients with iid or non-iid data.

\begin{itemize}
    \item Experiment a: 3 clients with IID data.
    \item Experiment b: 7 clients with data.
    \item Experiment c: 3 clients with Non-IID data.
    \item Experiment d: 7 clients with Non-IID data.
\end{itemize}

In the experiments of 3 client, we launch the client program to connect to the server on 3 Raspberry Pi devices, one of which has 4 GB of memory. And each client's training set have 20,000 samples. In experiments of 7 client, we launch the client program to connect to the server on 5 Raspberry Pi devices and 1 laptop (launched 2 processes). And each client's training set have 10,000 samples. The rest of the hardware configuration and software system have been mentioned in Section \ref{sec:setting}.

\begin{figure}[tbp]
\centering
\subfigure[Experiment a]{
\includegraphics[width=0.44\linewidth]{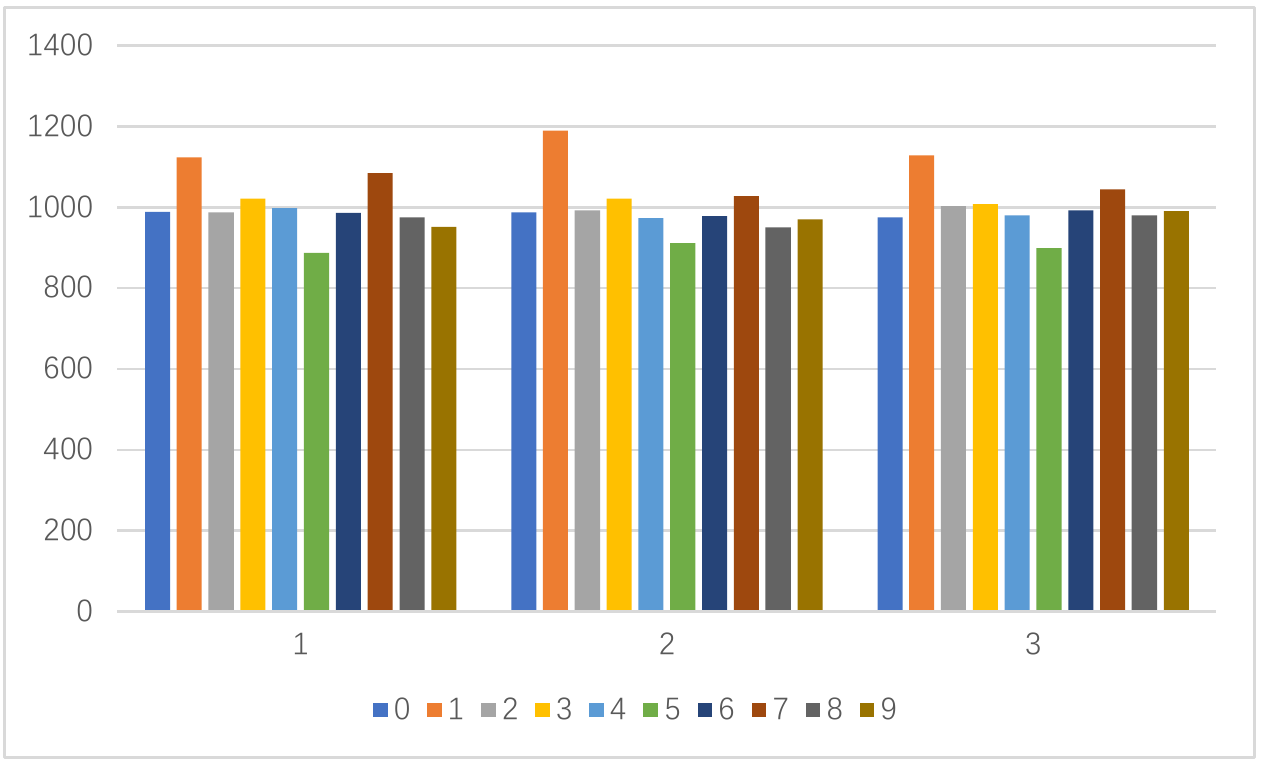}\vspace{4pt}
}
\quad
\subfigure[Experiment b]{
\includegraphics[width=0.44\linewidth]{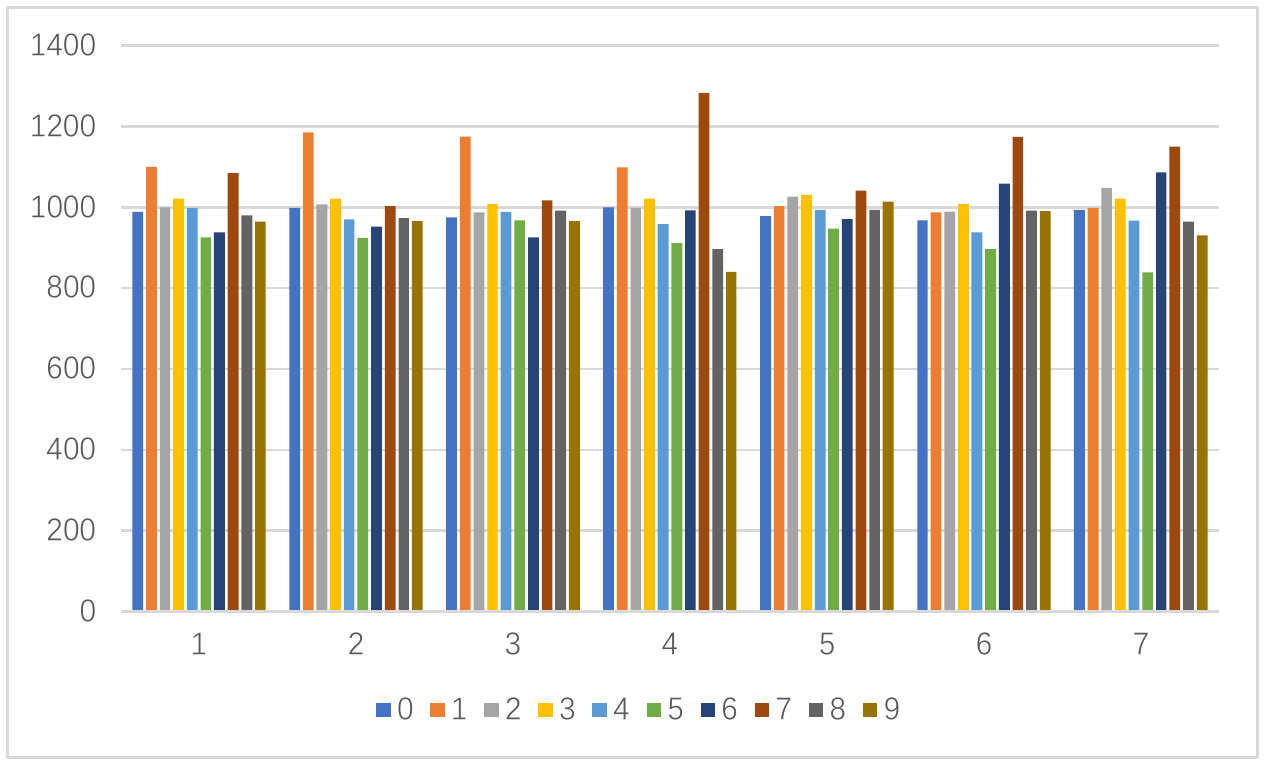}\vspace{4pt}
}
\quad
\subfigure[Experiment c]{
\includegraphics[width=0.44\linewidth]{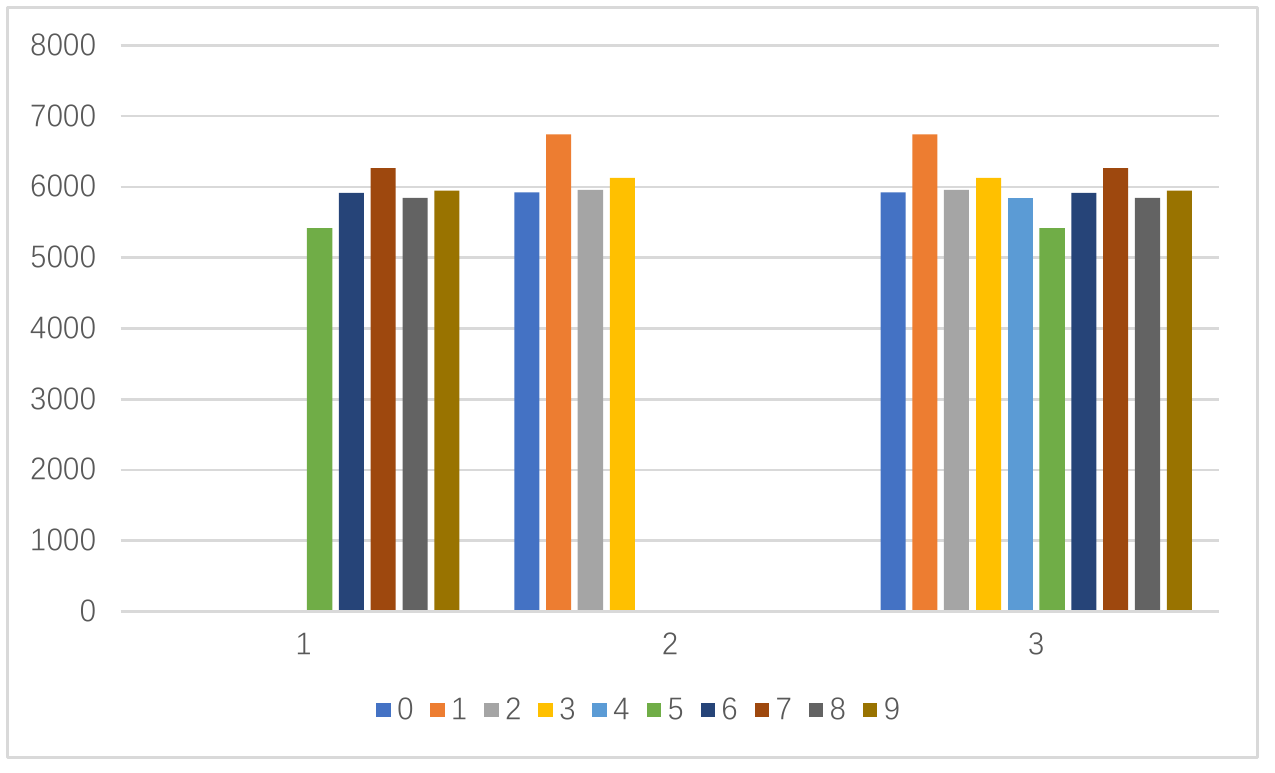}\vspace{4pt}
}
\quad
\subfigure[Experiment d]{
\includegraphics[width=0.44\linewidth]{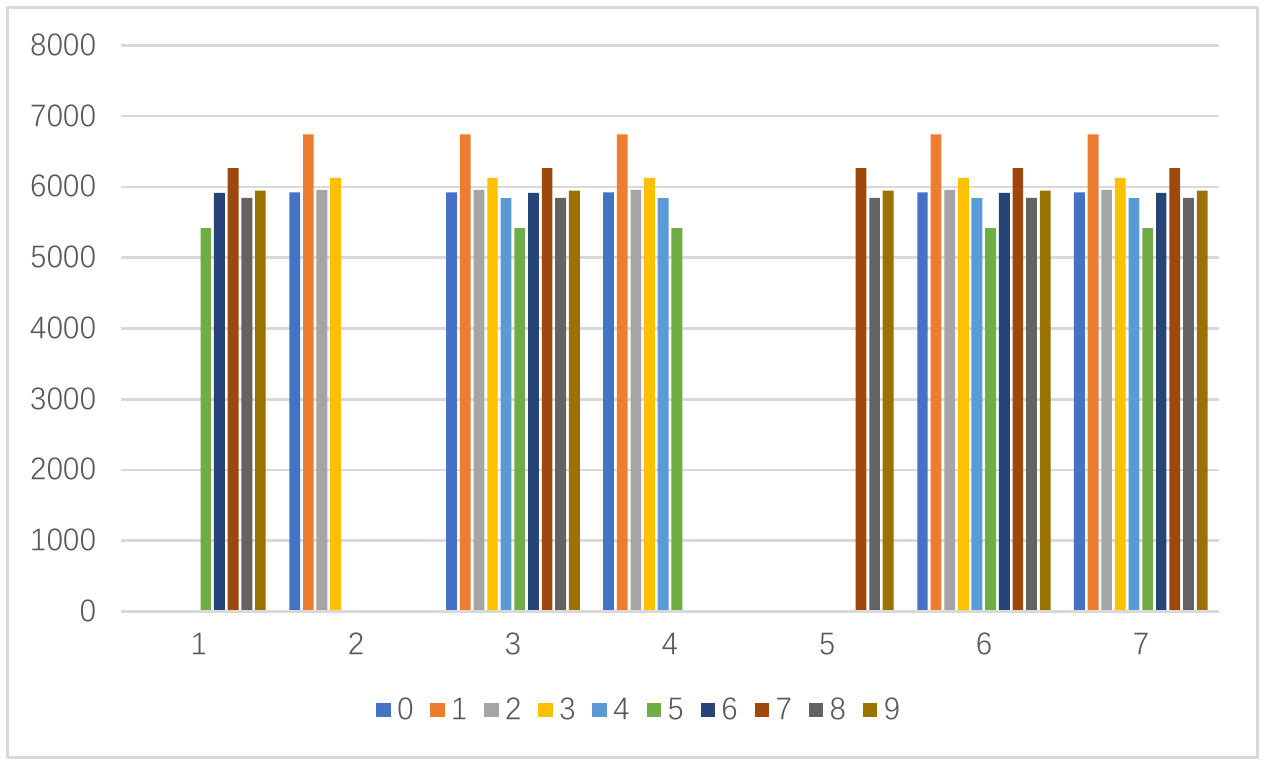}
}
\caption{Dataset distribution of clients in different experiments}
\label{data}
\end{figure}

The samples distribution of each client is shown in Fig. \ref{data}, indicating samples with 10 class labels (0-9) distributed with clients (1-3 or 1-7).

\begin{figure}[htbp]
\centering
\subfigure[Experiment a]{
\includegraphics[width=0.44\linewidth]{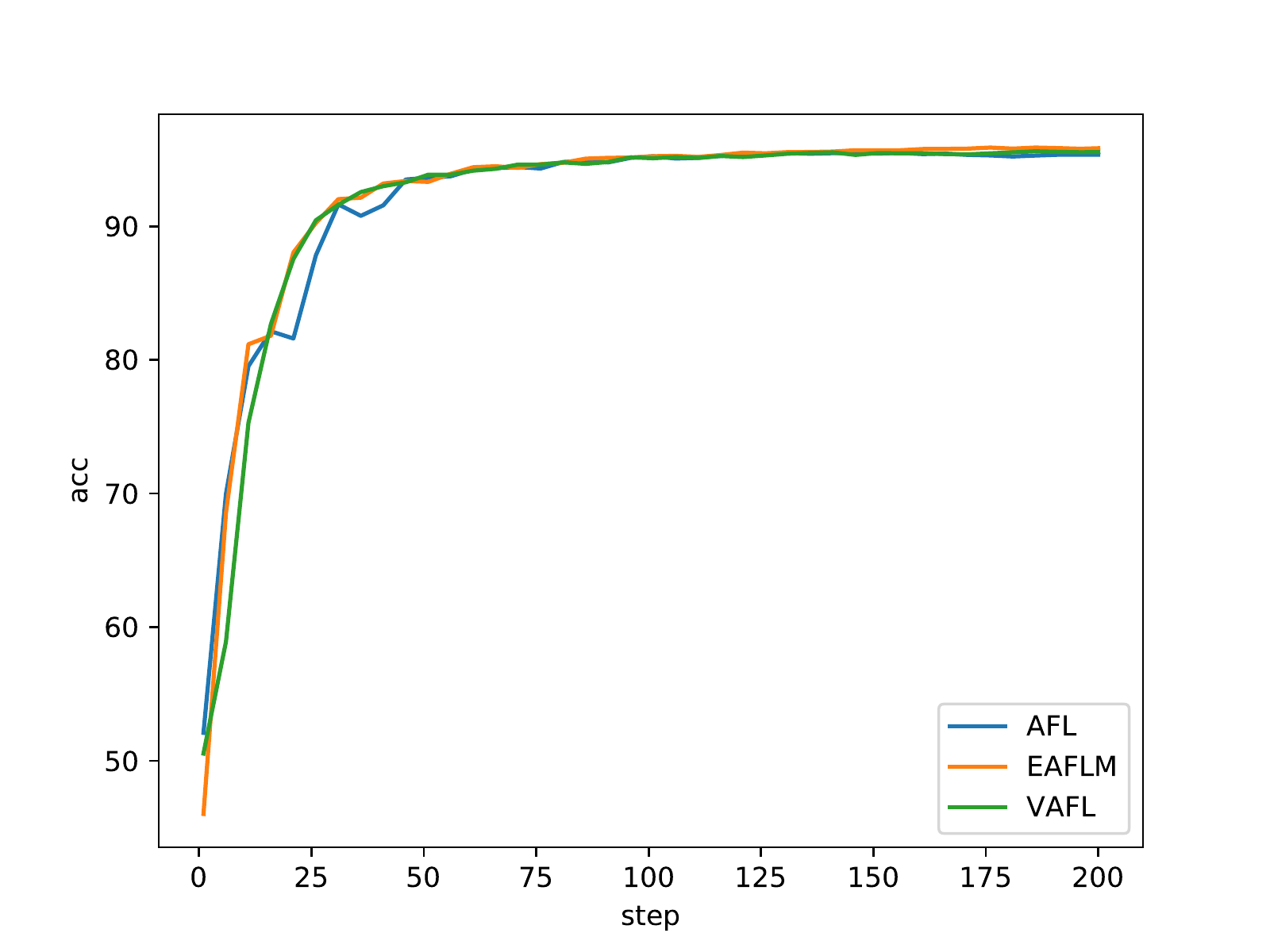}\vspace{4pt}
}
\quad
\subfigure[Experiment b]{
\includegraphics[width=0.44\linewidth]{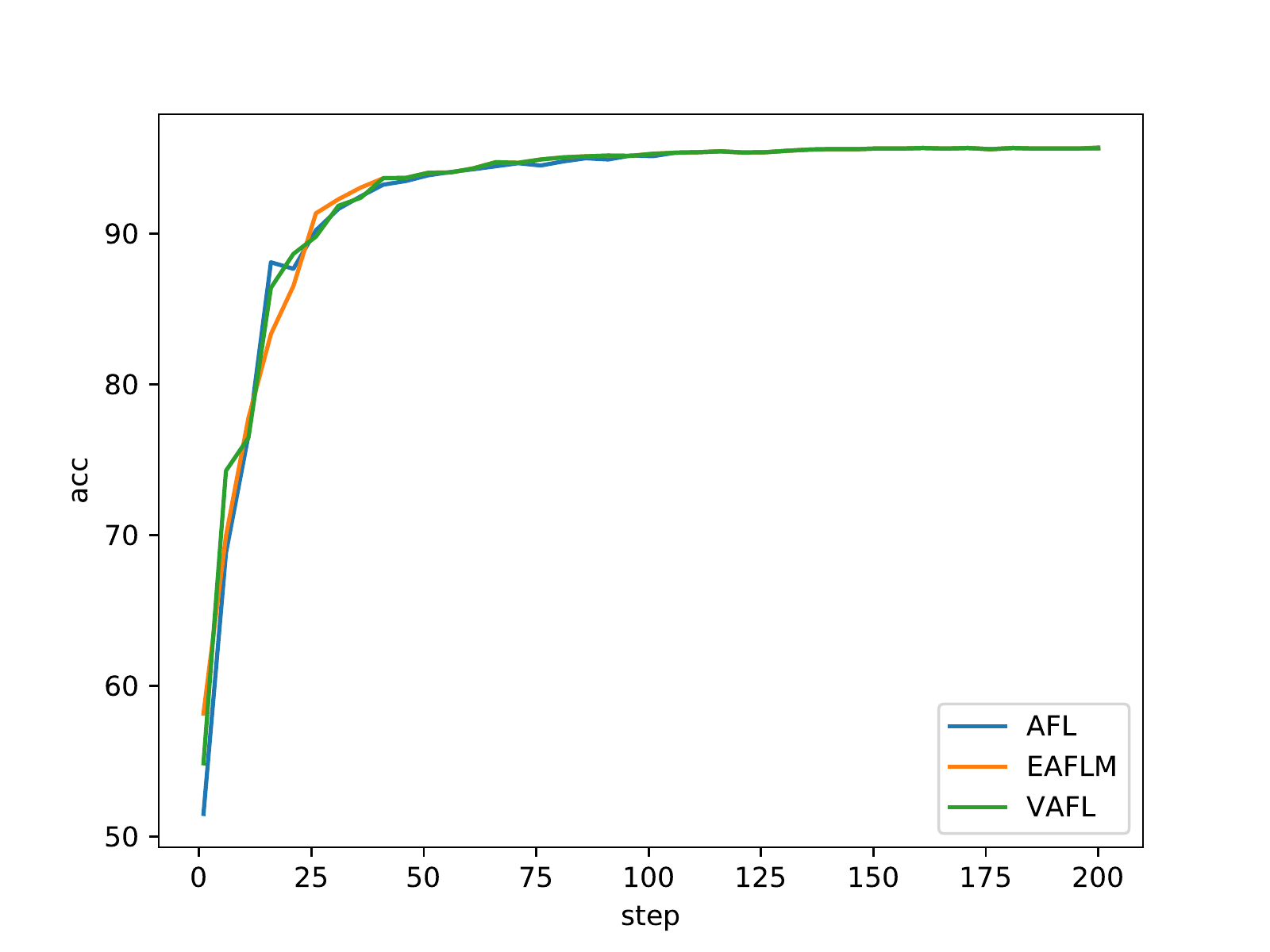}\vspace{4pt}
}
\quad
\subfigure[Experiment c]{
\includegraphics[width=0.44\linewidth]{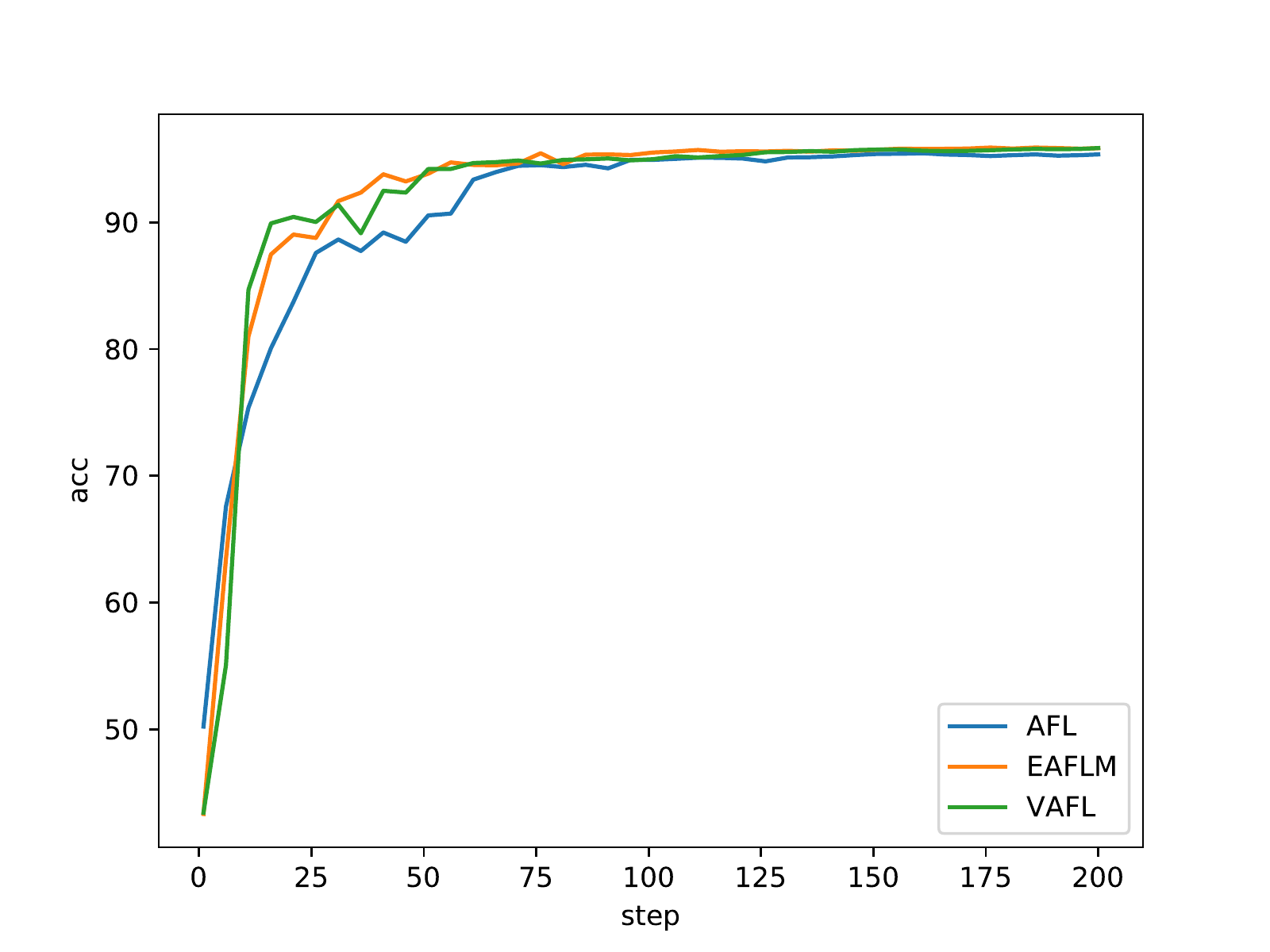}\vspace{4pt}
}
\quad
\subfigure[Experiment d]{
\includegraphics[width=0.44\linewidth]{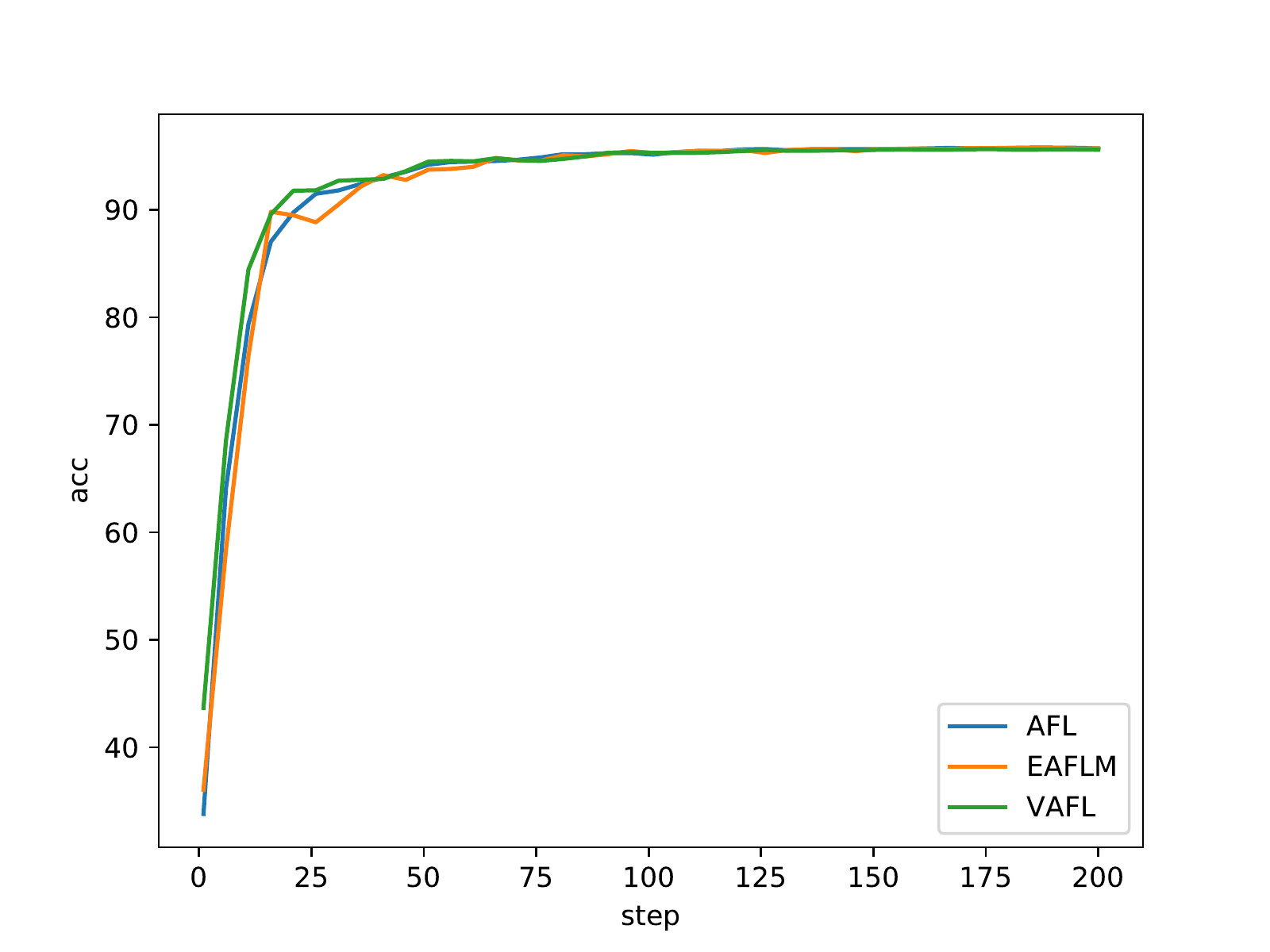}
}
\caption{$Acc$ of each algorithm in different experiments}
\label{Acc alg}
\end{figure}

The neural network we use has been mentioned in Section \ref{sec:setting} also. We perform image classification experiments to verify whether our federated learning asynchronous training process can converge and the effectiveness of VAFL. We compare the three methods, ordinary asynchronous training, VAFL method, and EAFLM. The $Acc$ of the federated model during the training period in four experiments are shown in Fig. \ref{Acc alg}.

We also compare the number of communications and the communication compression rate experienced by the three methods in training the model to achieve 94\% $Acc$. The experimental results are shown in Tab. \ref{tab3}.

\subsection{VAFL Performance in Multiple Experiments}
\begin{figure}[htbp]
\centering
\subfigure[Experiment a]{
\includegraphics[width=0.44\linewidth]{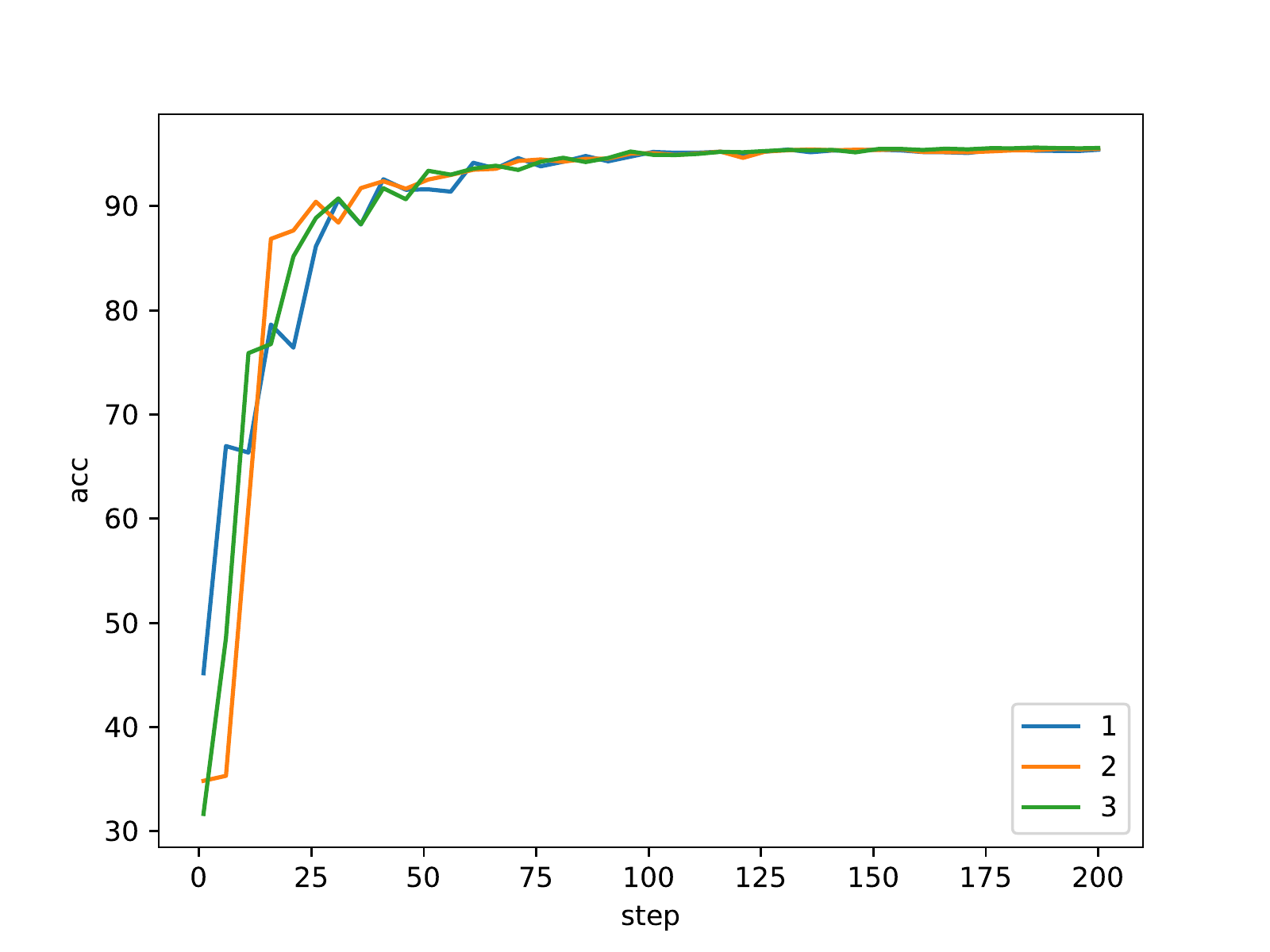}\vspace{4pt}
}
\quad
\subfigure[Experiment b]{
\includegraphics[width=0.44\linewidth]{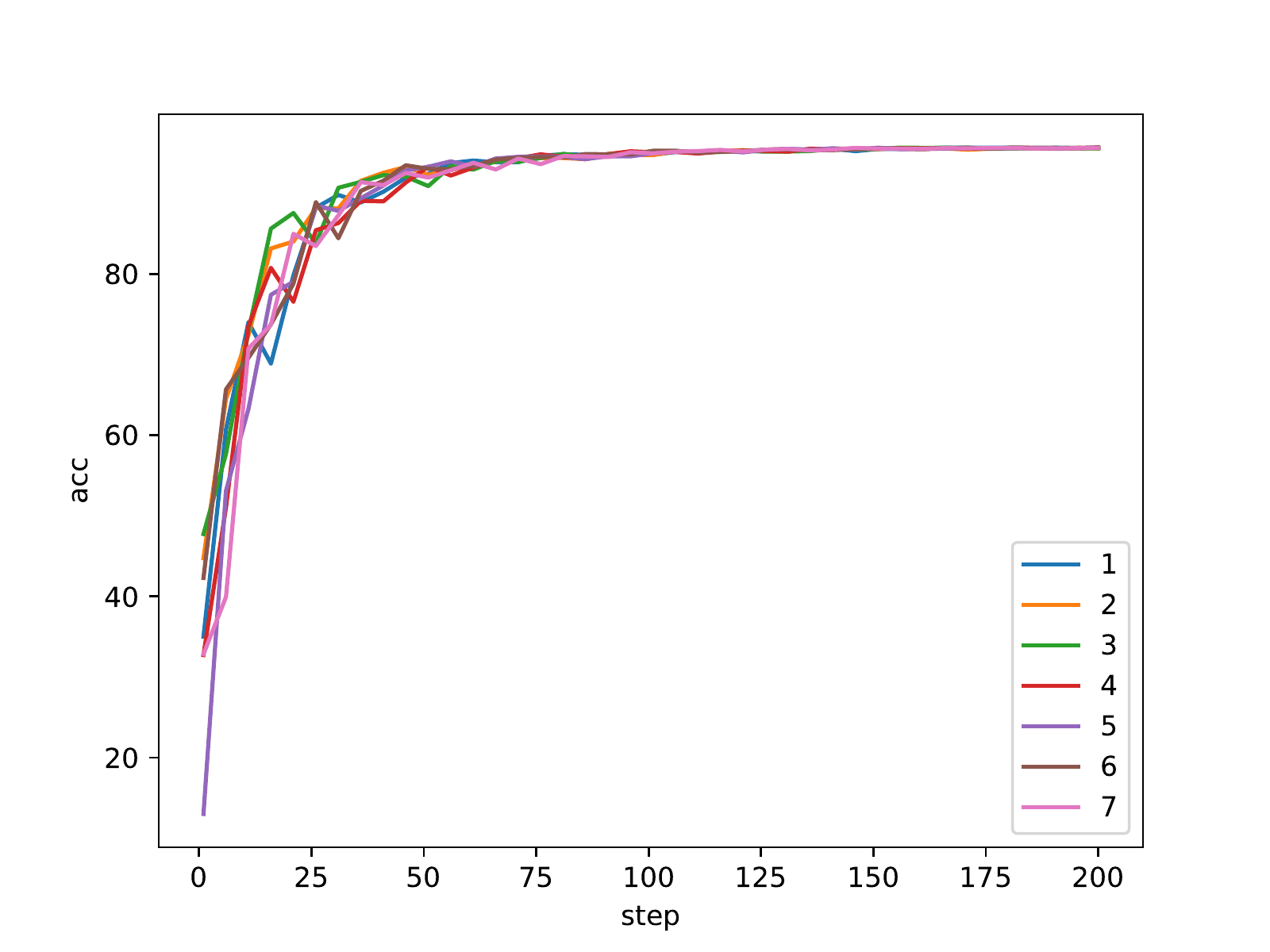}\vspace{4pt}
}
\quad
\subfigure[Experiment c]{
\includegraphics[width=0.44\linewidth]{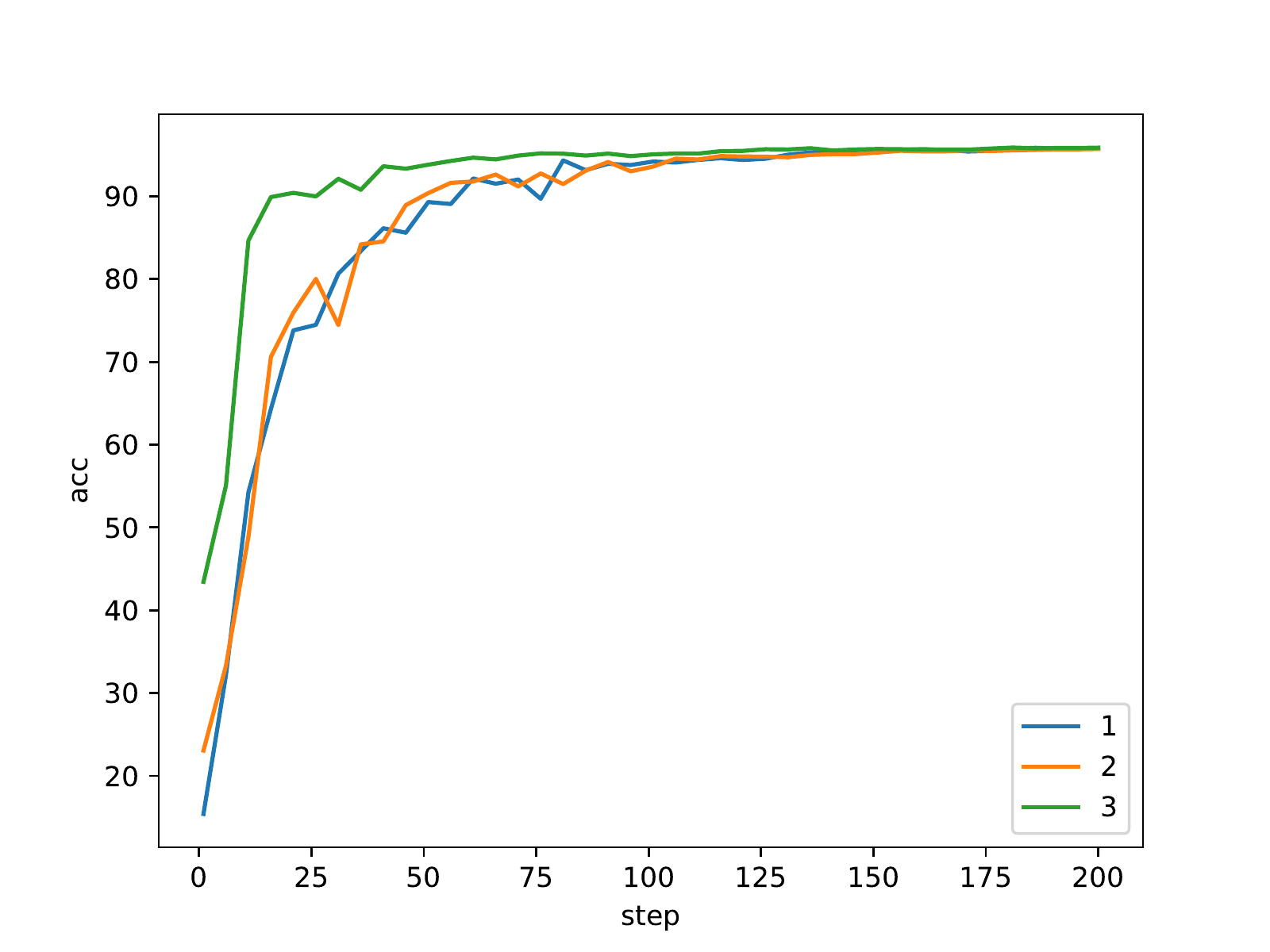}\vspace{4pt}
}
\quad
\subfigure[Experiment d]{
\includegraphics[width=0.44\linewidth]{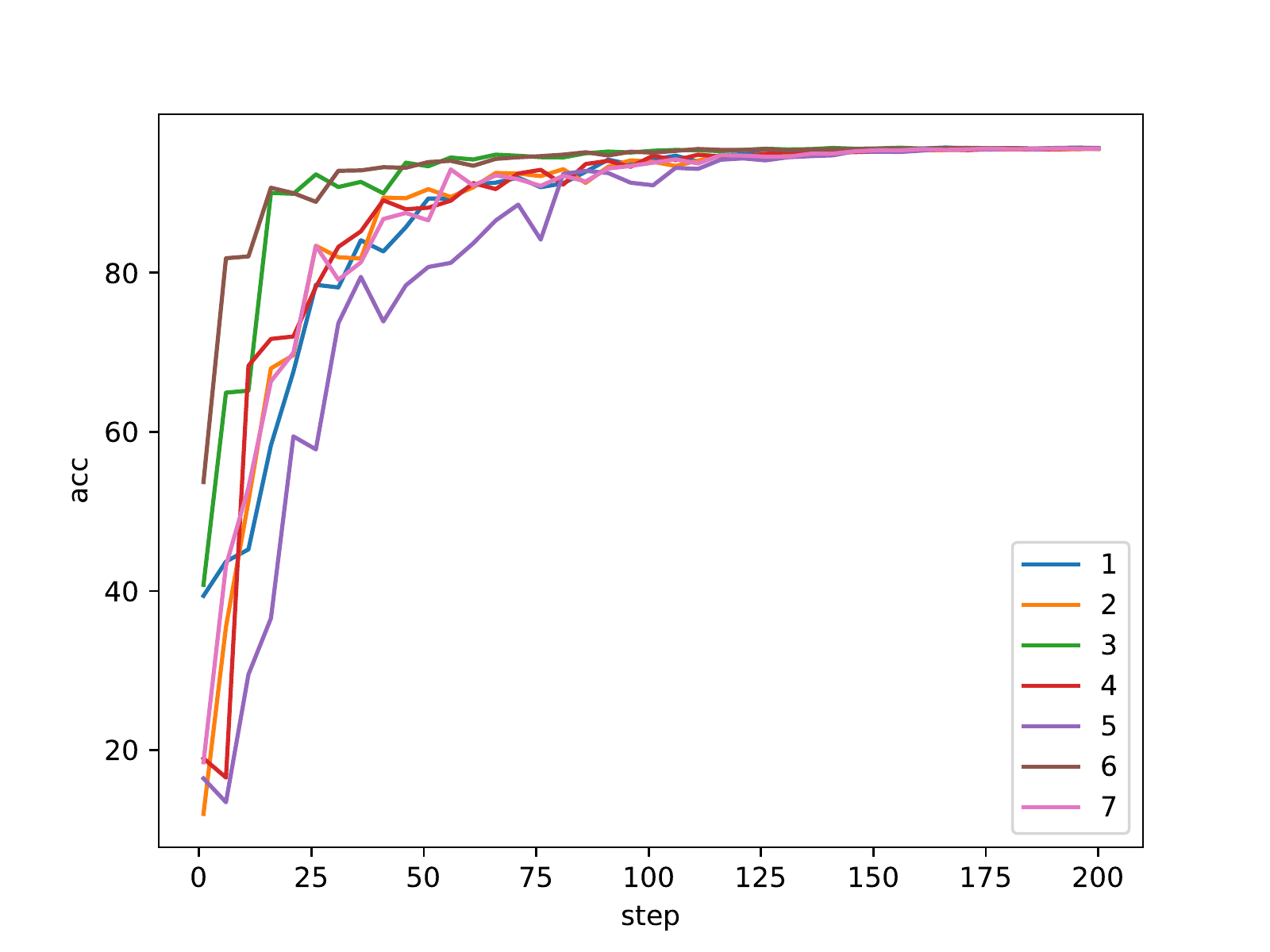}
}
\caption{$Acc$ of each client when executing VAFL}
\label{Acc clients}
\end{figure}

\begin{figure}[htbp]
\centerline{\includegraphics[width=2in]{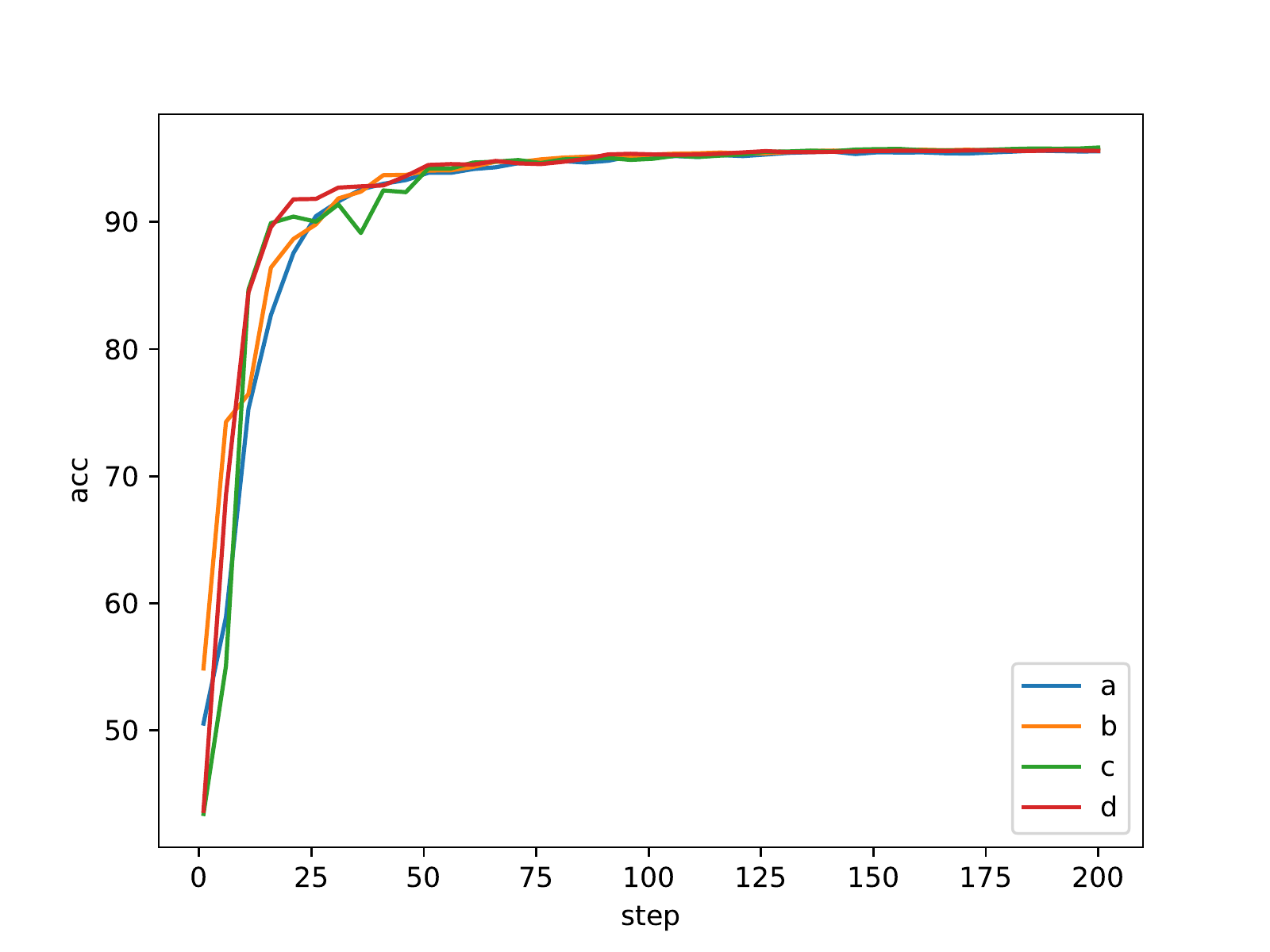}}
\caption{$Acc$ when executing VAFL in different experiments.}
\label{VAFL more}
\end{figure}

We also compared the performance of VAFL itself in different experiments. In the four experiments a, b, c, and d, the $Acc$ of each client when executing VAFL are shown in Fig. \ref{Acc clients}. And the $Acc$ when executing VAFL in different experiments is shown in the Fig. \ref{VAFL more}. The $\text{Communication times}$ and the $CCR$ are shown in the Tab. \ref{tab3}. 

As the number of clients increases and the imbalance in the distribution of the dataset intensifies, the better VAFL performs.

\begin{table}[htbp]
\caption{CCR and Communication times of different experiments}
\label{tab3}
\begin{center}
\begin{tabular}{cccc}
\hline
Experiment& Algorithm & Communication times & CCR             \\ \hline
\multirow{3}{*}{a} & AFL       & 39                  & 0               \\ \cline{2-4} 
                   & EAFLM     & 25                  & 0.3590          \\ \cline{2-4} 
                   & \textbf{VAFL}      & \textbf{28}         & \textbf{0.2821} \\ \hline
\multirow{3}{*}{b} & AFL       & 84                  & 0               \\ \cline{2-4} 
                   & EAFLM     & 45                  & 0.4643          \\ \cline{2-4} 
                   & \textbf{VAFL}      & \textbf{43}         & \textbf{0.4881} \\ \hline
\multirow{3}{*}{c} & AFL       & 45                  & 0               \\ \cline{2-4} 
                   & EAFLM     & 19                  & 0.5778          \\ \cline{2-4} 
                   & \textbf{VAFL}      & \textbf{22}         & \textbf{0.5111} \\ \hline
\multirow{3}{*}{d} & AFL       & 77                  & 0               \\ \cline{2-4} 
                   & EAFLM     & 35                  & 0.5455          \\ \cline{2-4} 
                   & \textbf{VAFL}      & \textbf{27}         & \textbf{0.6494} \\ \hline
\end{tabular}
\end{center}
\end{table}

\subsection{Summary}
Combining the results of the above experiments, it can be seen that compared with the basic asynchronous algorithm, VAFL enables the model to be trained faster, its $Acc$ can be improved faster in a short period. Also, it can reduce the communication time of 51.02\% and reach an average communication compression rate of 48.26\%. Compared with EAFLM, VAFL reduces the number of communications by about 3.23\% and improves the compression rate of communications by an average of about 0.82\%.

In summary, VAFL is able to compress the communication to a large extent, reduce the communication burden between the client and the server, and improve model convergence speed compared to the common asynchronous training methods. Compared with EAFLM, VAFL is able to compress the communication to a better level and reduce the communication times during training period. And VAFL performs better as the number of clients increases and the imbalance in the distribution of the data set intensifies.

\section{Conclusion}

In this paper. We proposed VAFL, a federated learning algorithm, focusing on the limitation and imbalance of the communication in asynchronous federated learning. The main idea of VAFL is that we expect to evaluate the communication values of each client to decide whether each one of them can participate in the optimization to compress communication values between server and clients. This study has designed a series of experiments aimed to identify the algorithm improving the efficiency of federated learning of edge devices compared with general algorithms and EAFLM, but also including some deficiencies.

VAFL enables the model to be trained faster and can reduce the communication time of 51.02\% and reach an average communication compression rate of 48.26\%. But VAFL relies heavily on honest clients because the calculation of $V_i$ depends on the clients themselves. If the clients report $V_i$ incorrectly, this will exacerbate the unfairness between the data and clients.

Future work can introduce other reference factors to make the computed communication value more reflective of the client's real situation and make sure the value is trustworthy, such as local training for a certain amount of time before computing the communication value or setting up some mechanism to make the computation of the client communication value less frequent.

The approach proposed in this paper has potential applications in the fields of Internet of Things\cite{chen2021design}, robotics\cite{liu2019lifelong}\cite{liu2019federated}\cite{liu2021peer}, and smart cities\cite{zheng2021applications}. In addition, the idea of value assessment may be applied in the incentive mechanism of Federated learning\cite{liu2020fedcm}\cite{yan2021fedcm}.


\bibliography{ref.bib}
\bibliographystyle{IEEEtran}

\end{document}